\documentclass{article}

\PassOptionsToPackage{numbers}{natbib}
\usepackage[preprint]{neurips_2022}
\usepackage[colorlinks,citecolor=blue, linkcolor=blue, urlcolor=blue]{hyperref}

\usepackage{amsmath}
\usepackage{multirow}
\usepackage[utf8]{inputenc} 
\usepackage[T1]{fontenc}    
\usepackage{hyperref}       
\usepackage{url}            
\usepackage{booktabs}       
\usepackage{amsfonts}       
\usepackage{nicefrac}       
\usepackage{microtype}      
\usepackage{xcolor}         
\usepackage{graphicx}
\usepackage{bbding}
\usepackage{pythonhighlight}
\usepackage{diagbox}

\title{Transformer-based dimensionality reduction}

\author{%
  Ruisheng Ran$^{1}$\thanks{$^{*}$Corresponding author}$^{*}$,  Tianyu Gao$^{1}$,  Yongdong Zhang$^{2}$ \\
  $^{1}$The College of Computer and Information Science, Chongqing Normal University, China \\
  $^{2}$The College of Computer Science, Chongqing University, China \\
}

\makeatletter
\def\thanks#1{\protected@xdef\@thanks{\@thanks
        \protect\footnotetext{#1}}}
\makeatother

\begin{document}

\maketitle

\begin{abstract}
Recently, Transformer is much popular and plays an important role in the fields of Machine Learning (ML), Natural Language Processing (NLP), and Computer Vision (CV), etc. In this paper, based on the Vision Transformer (ViT) model, a new dimensionality reduction (DR) model is proposed, named Transformer-DR. From data visualization, image reconstruction and face recognition, the representation ability of Transformer-DR after dimensionality reduction is studied, and it is compared with some representative DR methods to understand the difference between Transformer-DR and existing DR methods. The experimental results show that Transformer-DR is an effective dimensionality reduction method.
\end{abstract}

\section{Introduction}

In this information age, a large amount of data has been generated in various fields, including education, medical care, Internet, social media and business \cite{1}. These data are usually high-dimensional, heterogeneous, complex and massive \cite{2}, and they have different forms, such as text, digital image, voice signal and video.

Although the existing machine learning methods can also deal with high-dimensional data, they have some difficulties in dealing with such massive and high-dimensional data. The high dimension of the data will increase the complexity of the model, resulting in the slow training process of the model, low computational efficiency, and not conducive to the solution of the problem. This is the so-called "dimension disaster" problem \cite{3}\cite{4}. Some data contain thousands of features. If a model learns from many features, it will strongly rely on the data, resulting in overfitting and poor generalization \cite{5}. In addition, because the original data may contain redundant, insignificant or noisy information, the accuracy of the learned model will be poor \cite{3}.

Dimension reduction (DR) technology provides a method to reduce the dimension and preserve the effective features of data \cite{3}\cite{6}\cite{7}. It is usually used in the preprocessing stage of data before the machine learning model to improve the performance of the model. DR has the following advantages: 1) the lower data dimension reduces the computational complexity and improves the efficiency of the model training, 2) to reduce the complexity of the model and avoid the over-fitting problem, 3) it helps to eliminate redundant and insignificant features and noisy information, thus improving the accuracy of the model, 4) it is easy for data visualization and interpretability.

At present, researchers have proposed various Dimensionality Reduction methods \cite{3}\cite{6}\cite{7}\cite{8}. DR can be divided into linear and nonlinear methods. Representatives of linear methods are, Principal Component Analysis (PCA) \cite{9}, Linear Discriminant Analysis (LDA) \cite{10}, Locality Preserving Projections (LPP) \cite{11}, etc. Nonlinear methods include global structure-based methods such as Multi-Dimensional Scaling (MDS) \cite{12}, Isometric Mapping (ISOMAP) \cite{13}, and Kernel PCA (KPCA) \cite{14}, and local structure-based methods such as Laplacian Eigenmaps (LE) \cite{15}, Locally Linear Embedding (LLE) \cite{16}, t-Distributed Stochastic Neighbor Embedding (t-SNE) \cite{17}\cite{18}, Uniform manifold approximation and projection (UMAP) \cite{18}. According to whether the class label of the sample is adopted in the training process, it is divided into unsupervised method and supervised method. Most data dimensionality reduction methods are unsupervised methods, such as PCA, LPP, MDS, LE, LLE, etc. The representative supervised dimensionality reduction methods are LDA and ICA. According to whether the objective function contains a local optimal solution, it is divided into convex method and non-convex method. The objective function of the non-convex method contains the local optimal solution, and the representative method is Autoencoder \cite{19}; The convex method does not contain the local optimal solution, and most dimensionality reduction methods are convex methods. These dimensionality reduction algorithms have corresponding applications in different fields, and they each have their advantages when dealing with different data.

Some representatives of linear, nonlinear, unsupervised, supervised, and non-convex dimensionality reduction methods are introduced, including PCA, LPP, LLE, LDA, and Autoencoder. PCA is an unsupervised linear method, which is to find a linear transformation to map the data from the high-dimensional space to the low-dimensional space, and makes the variance of the low-dimensional data is as large as possible and makes the features as uncorrelated as possible \cite{9}. Different from the variance maximization theory of PCA, Linear Discriminant Analysis (LDA) is to perform low-dimensional projection on the original data. After the projection, the distances of the within-class data are as "close" as possible, and the distances of the between-class data are as "far" as possible, so that the data has good separability. LDA is a supervised, linear DR method \cite{10}. Laplacian Eigenmaps (LE) is a manifold-based DR method, which expects the adjacent points in the high-dimensional space to be as close as possible in the low-dimensional space, so that the neighborhood structure relationship of the original data can be maintained after DR \cite{7}. Locality Preserving Projections (LPP) \cite{11} is a linear approximation of LE, which is proposed to solve the "out-of-sample" problem \cite{20}. Locally Linear Embedding (LLE) is based on that each data point and its closest certain number of neighbors are viewed as a locally linear patch of the manifold and then reconstruct each data point from its neighbors in the corresponding subspace \cite{16}. 

Autoencoder (AE) is based on multi-layer neural network architecture, which is a nonlinear, unsupervised, non-convex DR method \cite{19}. AE consists of encoder and decoder, which are stacked by fully connected layers. The high-dimensional data is the input of the encoder for dimensionality reduction to get low-dimensional representation, and the low-dimensional data is the input of the decoder to reconstruct the original data, the network parameters are adjusted by backpropagation to reduce the error between the input data and the reconstructed data \cite{21}. Based on the design of AE, it can learn not only linear features but also nonlinear features. If the input images contain noise, we use Autoencoder for dimensionality reduction, the network can still extract useful features from the messy information. This is the Denoising Autoencoder (DAE), which trains a robust model \cite{22}. 

In recent years, as a new neural network structure, Transformer \cite{23} is much popular and plays an important role in the fields of Machine Learning. Transformer was first proposed to solve the problem of Natural Language Processing (NLP), and then it was also applied to the Computer Vision field, so the Vision Transformer (ViT) \cite{24} model was proposed for image classification. Based on ViT, we propose a dimension reduction model named Transformer-DR. On MNIST datasets and ImageNet datasets, we conduct experiments in data visualization, image reconstruction, and face recognition, investigate the dimensionality reduction ability of Transformer-DR, and compared it with some dimensionality reduction methods, so as to try to understand the difference between Transformer-DR and the state-of-the-art dimensionality reduction methods. The results show that Transformer-DR is an effective dimensionality reduction technique.

\section{Related Work}
Recurrent Neural Network (RNN) \cite{25} once occupied a dominant position in Natural Language Processing (NLP), but its parallel computing ability is severely limited due to the structure of RNN. Transformer was proposed by Vaswani et al. To solve the problem that RNN cannot be parallelized in NLP \cite{23}. Transformer is a deep neural network based on self-attention mechanism \cite{26}, which can capture the effect of semantic associations with long intervals. Transformer consists of encoder and decoder. The encoder consists of Multi-head Attention, Shortcut Connection \cite{27}, Layer Normalization \cite{28}, and Feed Forward \cite{29}. The decoder has one more Mutil-Head Attention and normalization layer than the encoder, which is used to receive the output of the encoder. Transformer can capture richer features with multi-head attention \cite{13}. It is precisely because of the advantages of Transformer that it can replace RNN in the field of NLP.

However, in the CV field, convolutional neural networks (CNN) \cite{30} still dominate. Inspired by the application of transformer in NLP, many researchers tried to apply transformer to the CV field. Recently, the Visual Transformer (ViT) model was proposed in Ref. \cite{24} And used in image classification, thereby migrating Transformer from the NLP field to the CV field for the first time. ViT only uses the encoder part of the Transformer. It divides the image into non-overlapping patches and then flattens them, embeds the position encoding to get patch embedding, and input the patch embedding to the stacked encoder to get the prediction result. When trained with enough data, ViT can perform well in image classification tasks. It shows that Transformer can also achieve good results in the CV field without CNN. After ViT, Transformer is widely used in various fields of computer vision.

\section{Methods}
The most important part of Transformer is the self-attention mechanism. With the self-attention mechanism, the correlation between Sequences (represented as the correlation between patches in ViT) can be found. Among these Sequences/patches, some information may be "redundant" or insignificant, so we can use the self-attention mechanism to calculate the correlation between patches, then remove redundant features and extract more useful features, so as to achieve the purpose of dimensionality reduction.

\begin{figure}
    \centering
    \includegraphics[width=\linewidth]{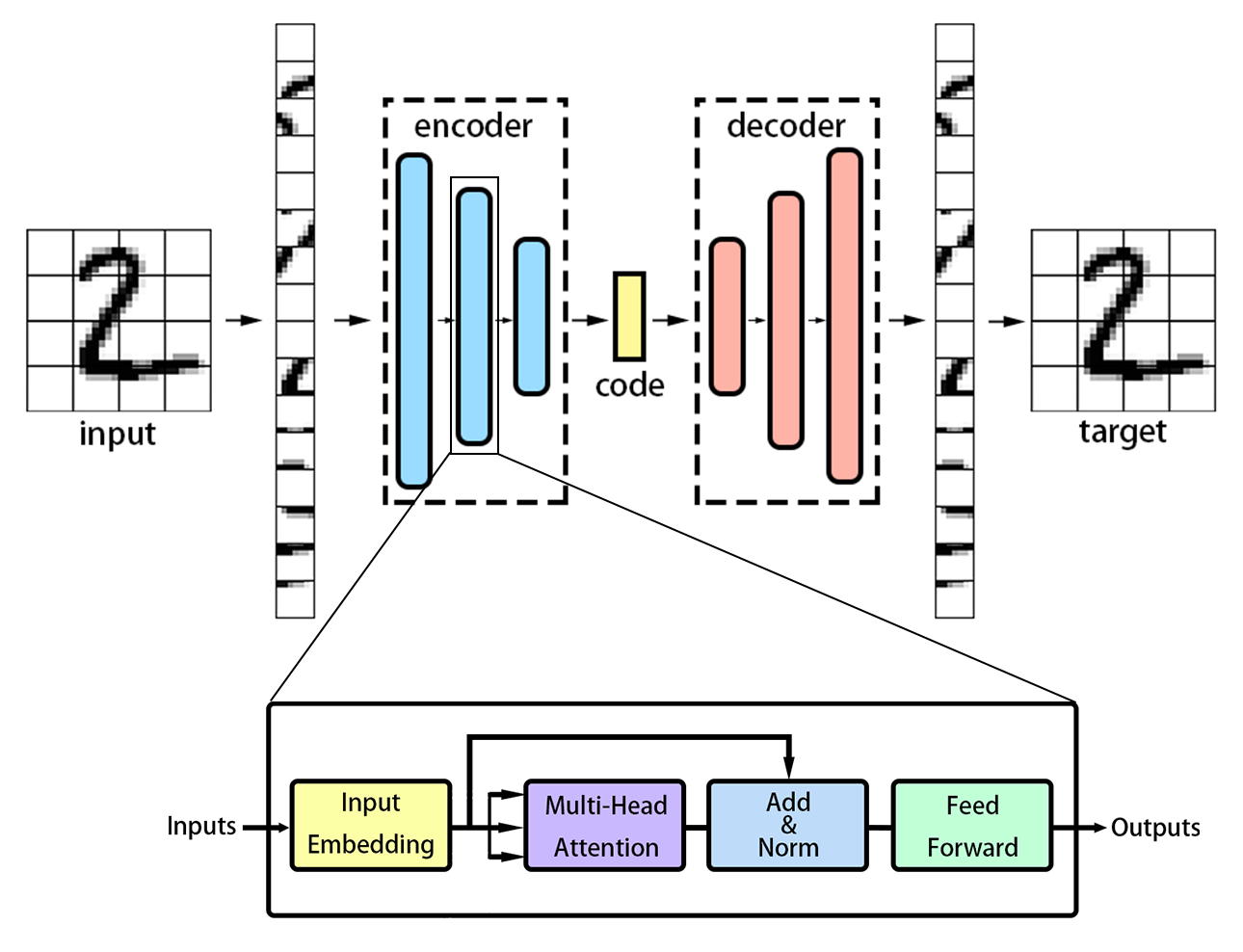}
    \caption{The Transformer-DR architecture.}
    \label{fig:1}
\end{figure}

We designed a network structure based on transformer, which is composed of encoder and decoder. Both the encoder and decoder are stacked by multiple Transformer blocks, as shown in Fig. \ref{fig:1}. Where, a Transformer block can be viewed as a component, after the input data passes through this component, the dimensionality of data will be reduced. Based on the network structure, the dimensionality of original data is gradually reduced by each Transformer block of the encoder, and then the original data is finally encoded into code. Then, the code is then gradually reconstructed into the original image by each Transformer block of the decoder. This model minimizes the error between the reconstructed image and the original image, so that the code from the encoder can best represent the original image.

For the 2D images, we divide the image into several non-overlapping patches, flatten the patches and add position encodings to get the patches embedding, then the patches embedding will be input into the encoder.

The encoder consists of multiple Transformer blocks stacked. In each Transformer block, the input data first passes through the Multi-Head Attention layer to calculate the attention mechanism and then passes through the Layer Normalization and Feed Forward layers to get the output data of the Transformer block. The Feed Forward layer of each Transformer block is expressed as:

\begin{equation}
 FFN(x)=GELU(xW_1+b_1)W_2+b_2.
\end{equation}
Where, $W_1\in\mathbb{R}^{{d_{input}}\times{d_{hidden}}}$, $b_1\in\mathbb{R}^{{d_{hidden}}}$, $W_2\in\mathbb{R}^{{d_{hidden}}\times{d_{output}}}$, $b_2\in\mathbb{R}^{{d_{output}}}$, $d_{hidden}$ is the dimension size of the hidden layer, $d_{input}$ is the size of data at each Transformer block entry, $d_{output}$ is the size of the output data after processed by the Feed Forward layer. And $d_{output}<d_{input}$, the dimensionality of data will be reduced in the Feed Forward layer of each Transformer block. In this way, after stacking multiple Transformer blocks, the original data will be embedded as a low-dimensional code.

The decoder is also a stack of multiple Transformer blocks. Different from the encoder, the Feed Forward layer of each Transformer block converts the data from low-dimensional to high-dimensional, and finally outputs the reconstructed image.

The structure of the encoder and decoder is symmetrical. That is, the number of Transformer blocks contained in the encoder and the decoder is the same. Secondly, at the corresponding positions of the encoder and the decoder, the number of nodes in the Feed Forward layer of each Transformer block is also symmetrical.

We use the Mean Square Error (MSE) as the loss function to update the network parameters by reducing the error between the original images and the reconstructed images.The formula of MSE is expressed as:
\begin{equation}
 MSE=\frac{1}{N}\sum_{i=1}^N||X_i-\hat{X_i}||^2
\end{equation}
where, $X_i$ is the original image, $\hat{X_i}$ is the reconstructed image of the model. In this way, the original data is encoded as code, and the dimensionality is reduced. The code can be regarded as an effective representation of the original data. Since the network structure uses Transformer for dimensionality reduction, we call it Transformer-based Dimensionality Reduction, which is abbreviated as Transformer-DR for convenience.

Compared with Autoencoder, the encoder and decoder of Transformer-DR are stacked by multiple Transformer blocks, while AE is stacked by restricted Boltzmann machines. From the experimental results in Section 4, after dimensionality reduction, Transformer-DR is better than Autoencoder for ordinary images. And in the case where the image is masked, due to the use of the self-attention mechanism, the quality of the reconstructed image after dimensionality reduction of Transformer-DR is better than that of AE.

Masked Autoencoder (MAE) is composed of encoder and decoder, it can reconstruct patches that have been randomly masked. Compared with MAE, MAE does not reduce the dimensionality of the data, and high-dimensional images still extract high-dimensional features. Secondly, MAE discards the masked image blocks at the input layer, only extracts the features of the non-masked image blocks, and then adds the mask information to form the image features \cite{31}.

\section{Experiments}
First, on the MNIST dataset, we use some representative dimensionality reduction methods to project the data of this dataset onto a two-dimensional plane for visualization. Secondly, we conduct image reconstruction experiments on MNIST and ImageNet datasets. In this experiment, we also randomly mask the original image before reconstruction, so as to study the image reconstruction ability of Transformer-DR model. Finally, we use the face dataset to do face recognition experiments after dimensionality reduction. In the experiments, Transformer-DR performed very well.

\begin{figure}
    \centering
    \includegraphics[width=\linewidth]{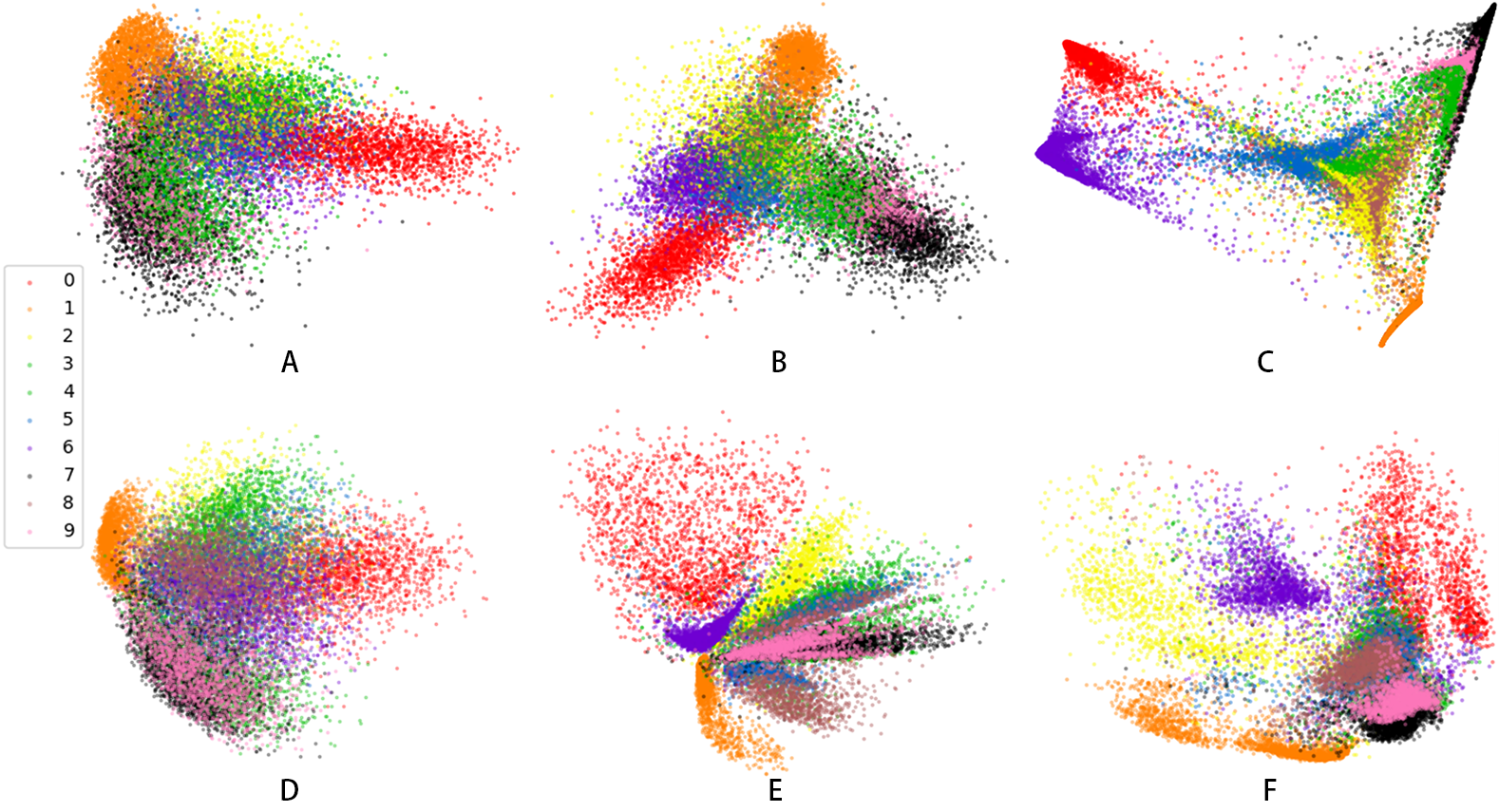}
    \caption{The comparison of data visualization results on the MNIST set.
(A) LPP. (B) LDA. (C) LLE. (D)PCA. (E) AE. (F) Transformer-DR.}
    \label{fig:3}
\end{figure}

\subsection{Data visualization}
In this section, we conduct data visualization experiments on the MNIST set. In the experiment, the data points are coded in 2-dimensional subspace with some representative dimensionality reduction methods, including PCA, LDA, LLE, LPP, AE and the proposed Transformer-DR. where, the network structure of AE is 784-512-256-128-64-2, and the network structure of Transformer-DR is (392$\times$2)-(256$\times$2)-(128$\times$2)-(64$\times$2)-(32$\times$2)-(1$\times$2). The visualization results are shown in Fig. \ref{fig:3}, where the different colors are used to represent different digit categories. As can be seen, compared with PCA, LDA, LLE, LPP, AE and Transformer-DR has more discrimination. The distribution effects of Transformer-DR and AE are comparable. 

\subsection{Image Reconstruction}
Here, we investigate the image reconstruction ability of Transformer- 
DR on the MNIST and ImageNet datasets, and compare it with PCA and AE.
\subsubsection{On the MNIST dataset}
\begin{figure}
    \centering
    \includegraphics[width=\linewidth]{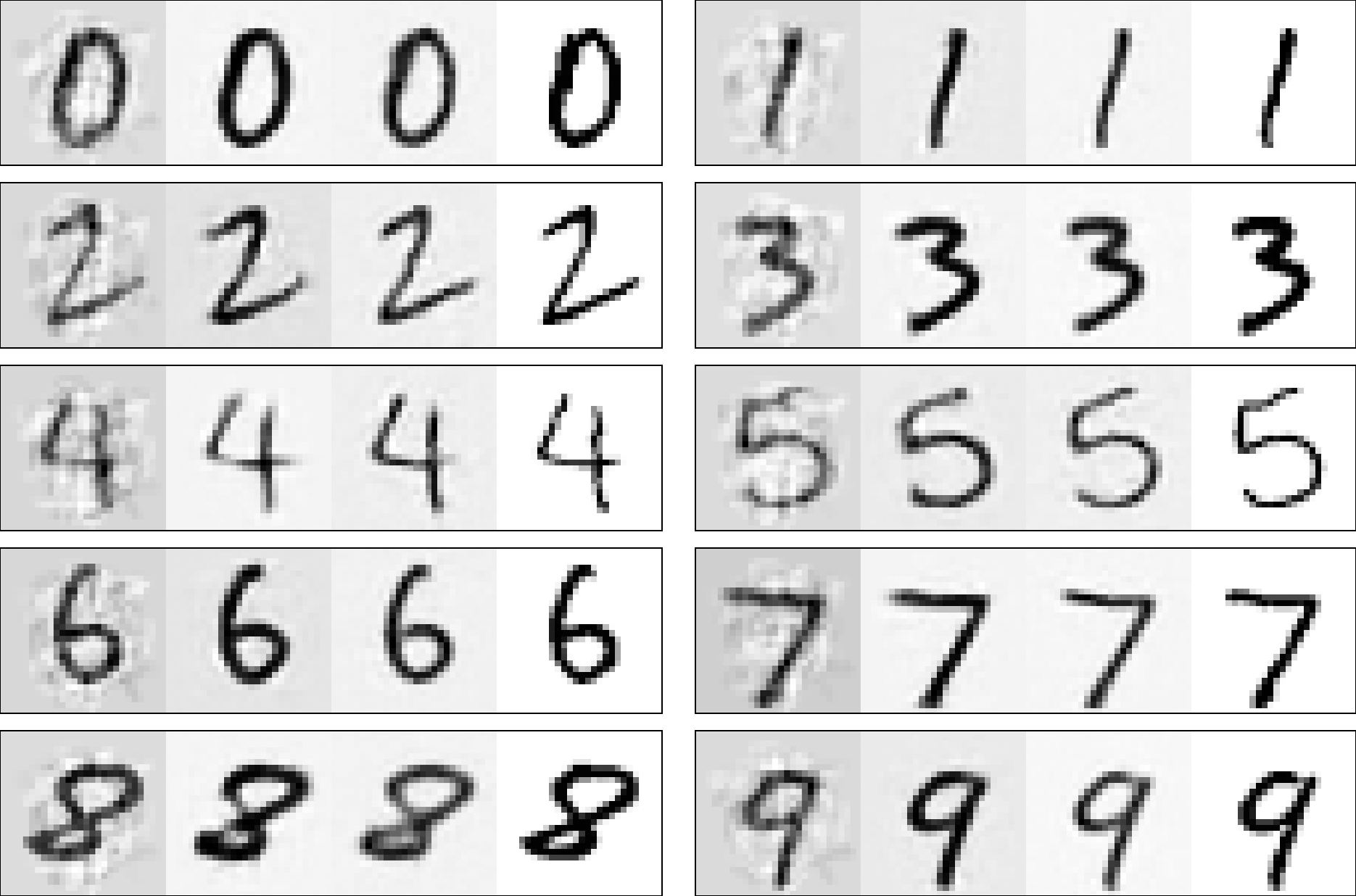}
    \caption{Every three columns are a group. For each group, from left to right: reconstructions of PCA; reconstructions of AE; reconstructions of Transformer-DR; the original images from the test dataset.}
    \label{fig:4}
\end{figure}

The MNIST dataset contains 60,000 training images and 10,000 test images. In the experiment, the size of the input original image is 28 $\times$ 28. For Transformer-DR, the 2D images are needed to divide into several non-overlapping patches before inputting into the encoder. So, in the network structure of Transformer-DR, the images are divided into 4 image blocks, each of which is 7 $\times$ 7 in size. The network structure of AE is 784-512-256-128-64-32. To ensure that the dimensions of each stage are equal to that of AE, the network structure of Transformer-DR is (196$\times$4)-(128$\times$4)-(64$\times$4)-(32$\times$4)-(16$\times$4)-(8$\times$4). Thus, the original images are all coded as 32-dimensional features. The experimental results are shown in Fig. \ref{fig:4}.

As can be seen from Fig. \ref{fig:4}, the reconstruction effects of Transformer-DR and AE are similar, but the reconstructed images of Transformer-DR are better than AE in some details. Furthermore, the reconstructed images of both are significantly better than those of PCA. In theory, the more image blocks are divided, the better the reconstruction of Transformer-DR, but in order to achieve the dimensionality reduction process equivalent to AE, we divide the image blocks as above in this experiment.

Due to the self-attention mechanism, Transformer can find correlations between image patches. In this way, the transformer-DR can also reconstruct the original image after dimensionality reduction if the image patches contain only a small amount of useful information. To verify this, we randomly masked 75\% of the image patches after dividing the image into 16 patches. After this processing, the image contains very little useful information. 

\begin{figure}
    \centering
    \includegraphics[width=\linewidth]{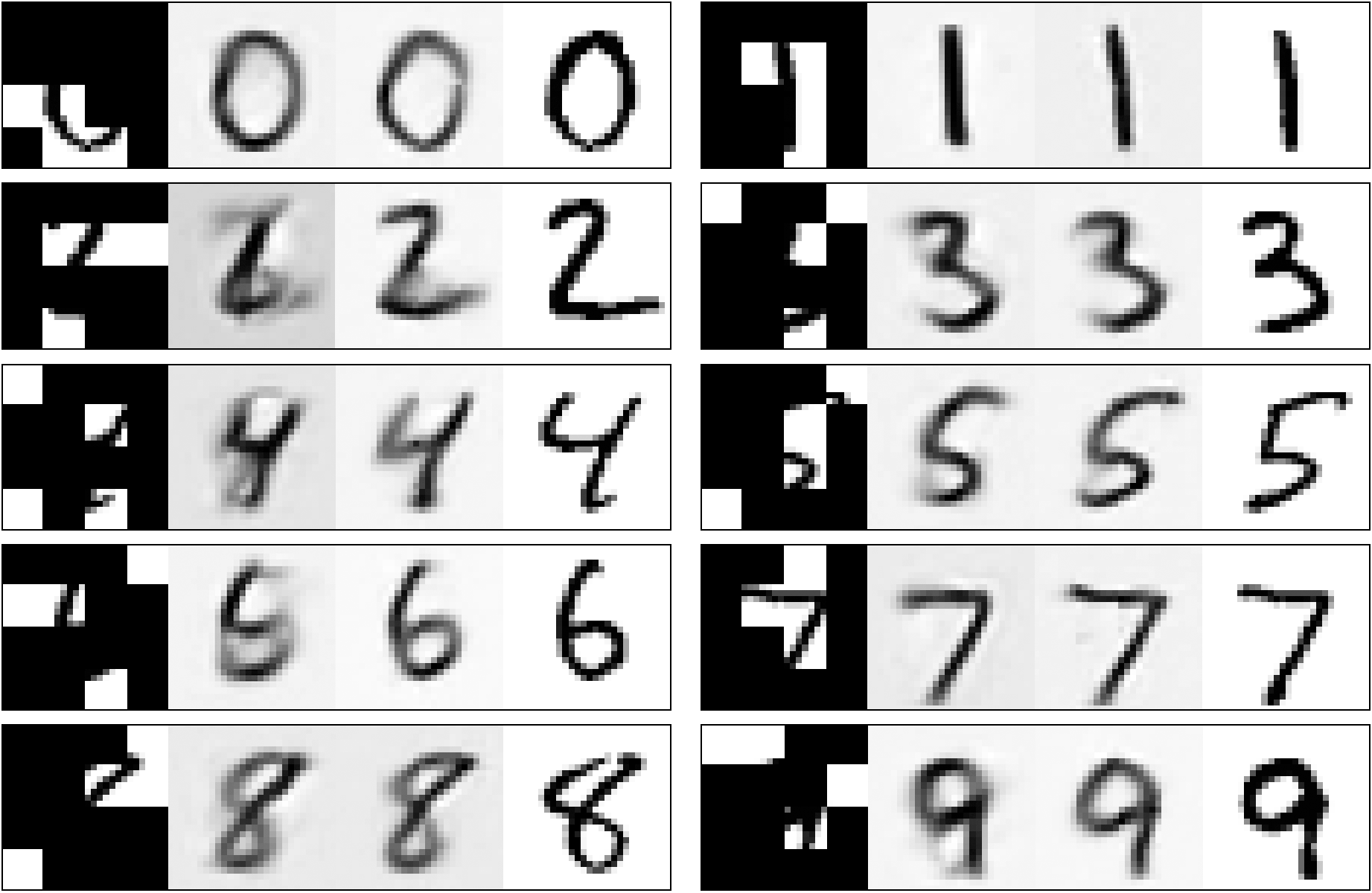}
    \caption{Every four columns are a group. For each group, from left to right: the masked images (masked ratio 75\%); reconstructions of AE; reconstructions of Transformer-DR; the original samples.}
    \label{fig:5}
\end{figure}

Then, we use AE and Transformer-DR for dimensionality reduction, the network structure used is the same as before, and the image is also dimensionally reduced to 32-dimensional features. The experimental results are shown in Fig. \ref{fig:5}. It can be seen from Fig. \ref{fig:5} that the transformer-DR can still reconstruct the dimensionally reduced images, and the reconstruction effects are better than that of AE.

\subsubsection{On the ImageNet dataset}
\begin{figure}
    \centering
    \includegraphics[width=\linewidth]{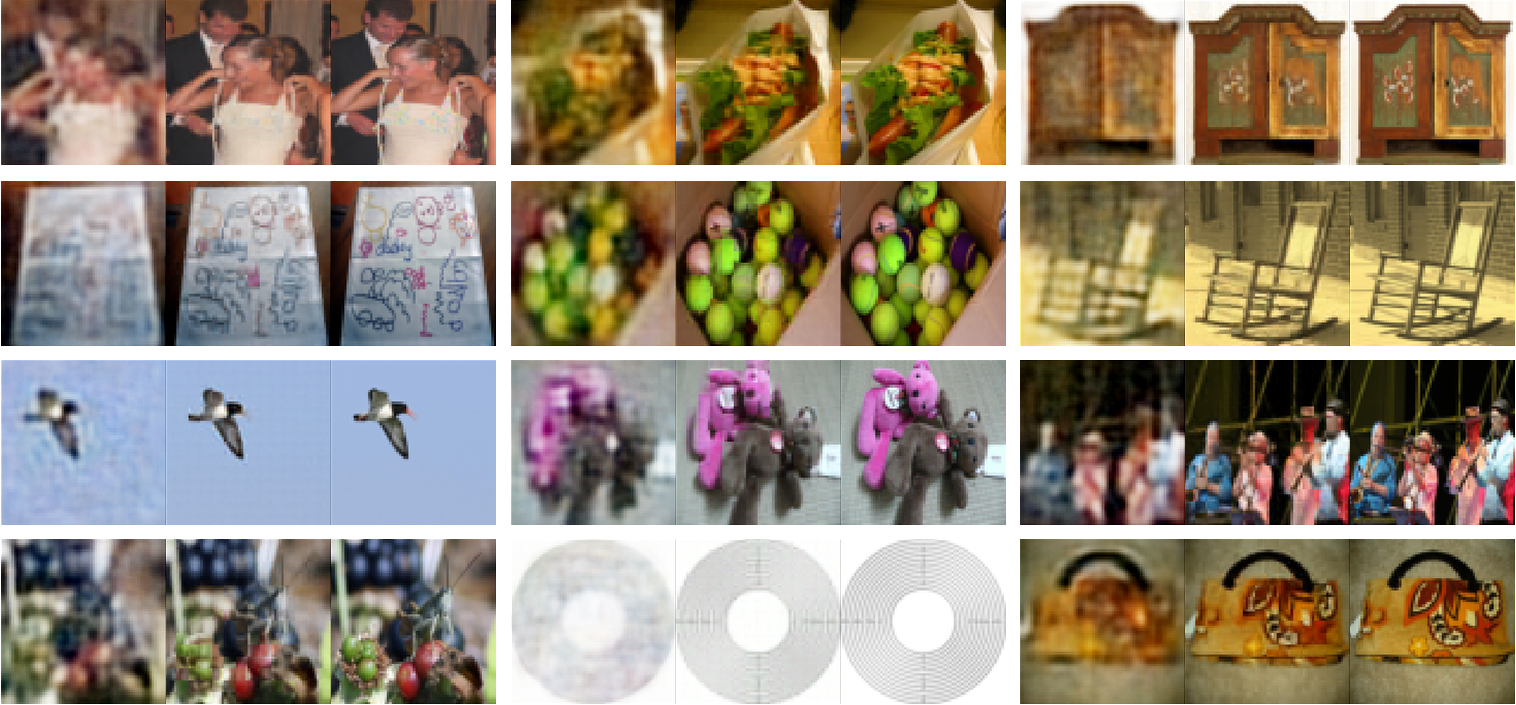}
    \caption{Every three columns are a group. For each group, from left to right: reconstructions of AE; reconstructions of Transformer-DR; the original images.}
    \label{fig:6}
\end{figure}

The ILSVRC-2012 ImageNet dataset \cite{32}  is larger and has more complex images. We make the image reconstruction experiments on this dataset. The network structure of AE and Transformer-DR is (70$\times$70$\times$3)-(3136$\times$3)-(1568$\times$3)-(784$\times$3). For Transformer-DR, the images are divided into 14 $\times$ 14 image patches, each of size 5 $\times$ 5. The experimental results are shown in Fig. \ref{fig:6}, which shows that the reconstructed images of Transformer-DR are almost the same as the original images and can handle the detail pixels well, while the reconstructed images of AE is blurry. And because AE all use fully connected layers, the training process will generate a large number of parameters, which makes AE unsuitable for training larger-scale data sets.

\begin{figure}
    \centering
    \includegraphics[width=\linewidth]{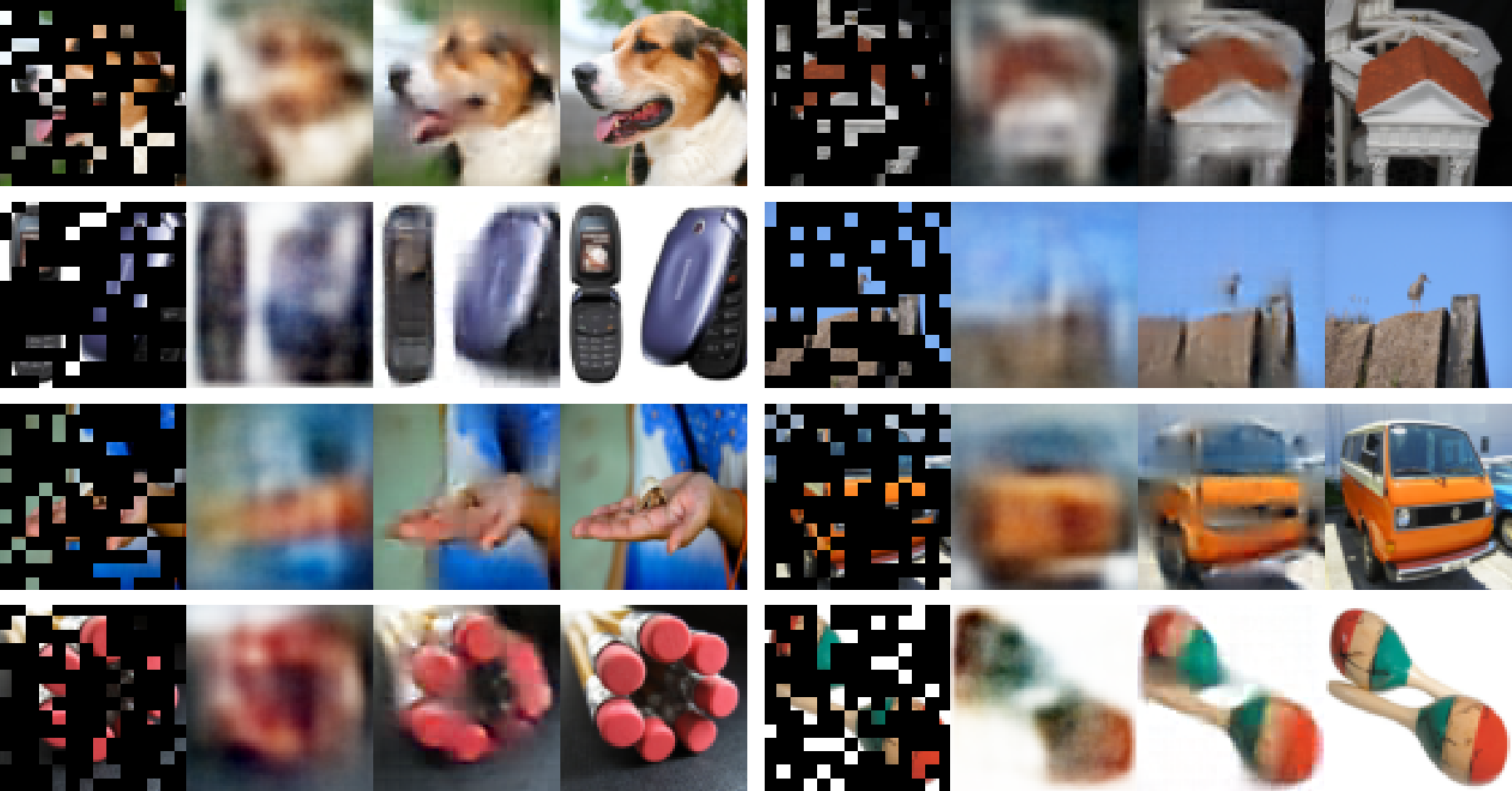}
    \caption{Every four columns are a group. For each group, from left to right: the image is masked by 75\%; reconstructions by AE; reconstructions by Transformer-DR; Random samples from the test data set.}
    \label{fig:7}
\end{figure}

Then, the experiment to reconstruct the original images from the masked images is also performed to explore the ability to extract robust features in more complex images. The experimental environment and variables are the same as in the above experiments, the only change is that 75\% of the patches are masked. The experimental results show that the reconstructed images of AE are so blurry that it is difficult for the eye to judge its original situation, while the reconstructed image of Transformer-DR can almost predict the mask area of the original image, see Fig. \ref{fig:7}.

\begin{figure}
    \centering
    \includegraphics[width=0.8\linewidth]{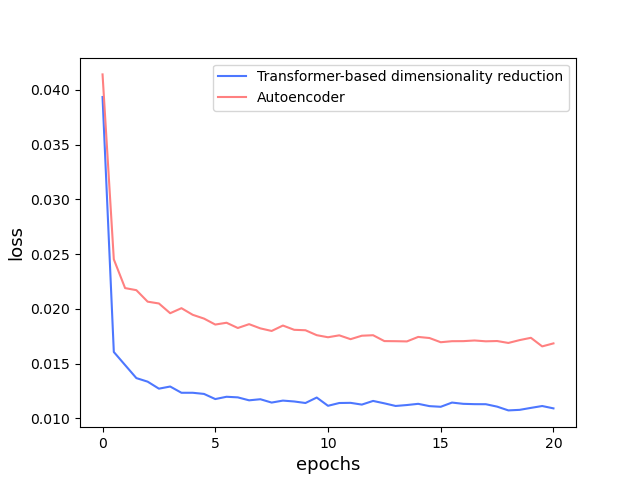}
    \caption{The loss value curves of both models at training stage.}
    \label{fig:8}
\end{figure}

Fig. \ref{fig:8} shows the loss values of Transformer-DR and AE trained for 20 epochs in this experiment. As can be seen from Fig. \ref{fig:8}, the loss values for both models drop significantly within 5 epochs and then slowly drop off. In the end,, the loss value of Transformer-DR is around 0.011, and the loss value of AE is around 0.017. Therefore, the training effect of Transformer-DR is better than that of AE.

\subsection{Face Recognition}
In this section, we will use the features after dimensionality reduction with Transformer-DR for face recognition. In the experiment, besides the Transformer-DR model for face recognition, we also design another dimensionality reduction structure, Transformer-DRr, for face recognition, it is trained with two loss functions, one for image reconstruction and one for face recognition. Face Transformer \cite{33} is used to compare with Transformer-DR and Transformer-DRr. Face Transformer is based on ViT, it introduces ViT to the field of face recognition and gets good recognition results. In addition, the ViTs model, which is based on the ViT model and using the sliding window to generate overlapping image patches, is also used for face recognition. And the performance of ViTs is slightly better than that of ViT, so our network is implemented based on ViTs. CosFace \cite{35} is selected as the loss function to improve the recognition performance.

The MS-Celeb-1M dataset \cite{34} is used as the training dataset, which contains 100,000 celebrities and nearly 10 million face pictures. The trained models are validated on LFW \cite{36}, SLLFW \cite{37}, CALFW \cite{38}, CPLFW \cite{39}, TALFW \cite{40}, CFP-FP \cite{41}, AgeDB-30 \cite{42} datasets. The experimental results are shown in Table \ref{table1}. Before training and validation, the face images are aligned to 112 $\times$ 112. The input dimensionality of Transformer-DR and Transformer-DRr is 37632 (112 $\times$ 112 $\times$ 3), after dimensionality reduction, the feature size is 12544. ViT and ViTs do not reduce the dimensionality of the original images, so the original images are stretched to 37632 dimensions, and the features are still 37632 dimensions after the processing of ViT and ViTs.

The face recognition results are shown in Table \ref{table1}. Where, the face recognition results of Face Transformer are quoted from Ref. \cite{35}. As can be seen from Table \ref{table1}, the recognition rates of Transformer-DR and Transformer-DRr are slightly lower than that of ViT and ViTs, but they are still high and close to that of ViT and ViTs. Note that Transformer-DR and Transformer-DRr reduce the original images to 12544-dimensional features for face recognition, while ViT and ViTs do not reduce the dimensionality of the original images and uses the 37256-dimensional features for face recognition. This shows that, after Transformer-DR dimensionality reduction, although the recognition rates have some slight losses, the high recognition rates can still be achieved. In other words, Face Transformer does not need such a high dimensional data for face recognition, and the high-dimensional original images may contain a lot of redundant information. And Transformer-DR can effectively remove some of the redundant data in the original images. The recognition ratio of the other dimensionality reduction methods in the LFW dataset is as follows, EigenFace \cite{43} is 60.85\%, LE is 82\%+ \cite{44} , DFD is 80.02\% $\pm$ 0.5\% \cite{45}, PCANet is 86.28\% $\pm$ 1.14\% \cite{46}, Their results are not as good as Transformer-DR.

Transformer-DRr has a similar recognition results with Transformer-DR, but it can not only perform face recognition but also image reconstruction.It is worth noting that the accuracy of transformer DRr on LFW dataset can reach 99.75\% $\pm$ 0.25\%, which shows that Transformer-DRr has reduced the dimension of the face image, but the loss of information is small, and its recognition results are very close to those of ViT and ViT.

\begin{table}[t]
	\centering
	\renewcommand{\arraystretch}{2}
	\caption{The recognition accuracy}
	\label{table1}
	\scalebox{0.5}{}
		\resizebox{\columnwidth}{!}
	{	
		\begin{tabular}{cccccccc}
			\hline
			Method       & LFW  & SLLFW  & CALFW  & CPLFW & TALFW & CFP-FP & AgeDB-30 \\
			\hline
			EigenFace   & 60.85  \\
			LE          & 82+ \\
			DFD        & 80.02 $\pm$ 0.5 \\
			PCANet     & 86.28 $\pm$ 1.14 \\
			ViT       & 99.83 & 99.53 & 95.92 & 92.55 & 74.87 & 96.19 & 97.82 \\
			ViTs       & 99.80 & 99.55 & 96.18 & 93.08 & 70.13 & 96.77 & 98.05 \\
			\textbf{Transformer-DR}       & \textbf{99.53 $\pm$ 0.31} & \textbf{98.82 $\pm$ 0.47} & \textbf{95.33 $\pm$ 1.03} & \textbf{90.25 $\pm$ 1.09} & \textbf{63.40 $\pm$ 1.80} & \textbf{93.77 $\pm$ 1.25} & \textbf{96.37 $\pm$ 0.97} \\
			\textbf{Transformer-DRr}       & \textbf{99.75 $\pm$ 0.25} & \textbf{98.90 $\pm$ 0.50} & \textbf{95.43 $\pm$ 1.11} & \textbf{90.23 $\pm$ 1.31} & \textbf{62.63 $\pm$ 1.31} & \textbf{93.64 $\pm$ 1.25} & \textbf{95.93 $\pm$ 0.88} \\
			\hline
		\end{tabular}
	}
\end{table} 

\section{Conclusion}
In ViT model, the image is divided into many patches and some information of these patches may be "redundant" or insignificant.  Based the self-attention mechanism of Transformer, we can find the correlation between patches and then remove redundant information and extract useful features. Then, a new dimensionality reduction (DR) model named Transformer-DR is proposed. The model consists of an encoder and a decoder, which are stacked by multiple Transformer blocks. Based on the model, the original image can be encoded into low dimensional code, thus achieving the goal of dimensionality reduction. We do experiments from data visualization, image reconstruction and face recognition, and the results show that Transformer DR method is superior to existing data dimension reduction methods.

\bibliographystyle{nips}
\bibliography{ref}
\end{document}